# A Quantum-Powered Photorealistic Rendering

Yuan-Fu Yang
National Yang Ming Chiao Tung University
yfyangd@gmail.com

Min Sun
National Tsing Hua University
summin@ee.nthu.edu.tw

## Abstract

*Achieving photorealistic rendering of real-world scenes poses a significant challenge with diverse applications, including mixed reality and virtual reality. Neural networks, extensively explored in solving differential equations, have previously been introduced as implicit representations for photorealistic rendering. However, achieving realism through traditional computing methods is arduous due to the time-consuming optical ray tracing, as it necessitates extensive numerical integration of color, transparency, and opacity values for each sampling point during the rendering process. In this paper, we introduce Quantum Radiance Fields (QRF), which incorporate quantum circuits, quantum activation functions, and quantum volume rendering to represent scenes implicitly. Our results demonstrate that QRF effectively confronts the computational challenges associated with extensive numerical integration by harnessing the parallelism capabilities of quantum computing. Furthermore, current neural networks struggle with capturing fine signal details and accurately modeling high-frequency information and higher-order derivatives. Quantum computing's higher order of nonlinearity provides a distinct advantage in this context. Consequently, QRF leverages two key strengths of quantum computing: highly non-linear processing and extensive parallelism, making it a potent tool for achieving photorealistic rendering of real-world scenes.*

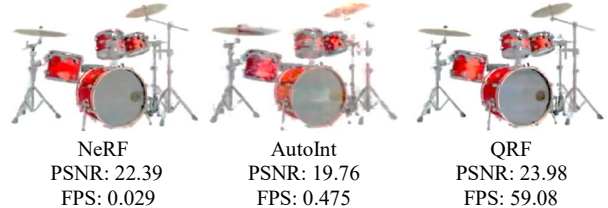

| NeRF | AutoInt | QRF |
|---|---|---|
| PSNR: 22.39 | PSNR: 19.76 | PSNR: 23.98 |
| FPS: 0.029 | FPS: 0.475 | FPS: 59.08 |

Figure 1: The results of volume rendering for Drums after training for 50k iterations. In comparison to the NeRF baseline [13] and AutoInt [17], QRF exhibit improved rendering efficiency and enhanced rendering quality.

## 1. Introduction

Traditional 3D computer vision pipelines depend on multi-view stereo algorithms for estimating sparse point clouds, camera poses, and textured meshes from 2D input views. However, re-rendering these scene representations often falls short of achieving photorealistic image quality. In stark contrast, implicit scene representations offer a substantial leap in rendering quality and can be directly supervised using 3D data through neural networks. Nonetheless, current neural networks grapple with capturing fine signal details and accurately modeling high-frequency information and higher-order derivatives, even when subjected to extensive supervision. Furthermore, achieving realistic rendering of real-world scenes using conventional computer graphics techniques is an imposing challenge, as it involves the intricate task of capturing nuanced appearance and geometric models. In practice, existing methods frequently produce blurry renderings due to limitations in network capacity. Generating high-resolution imagery from these representations often requires time-consuming optical ray tracing.

Recent strides in quantum computing have substantially enriched the realm of artificial neural networks. Quantum Neural Networks (QNN), a class of quantum algorithms harnessing qubits to construct adaptive neural networks, have played a pivotal role in this advancement. Quantum computing holds the promise of achieving exponential speed enhancements by executing massively parallel computations on superimposed quantum states. QNN tailored for implicit representations exhibit promise in surmounting the computational challenges that conventional techniques encounter when executed on classical computers. In this study, we employ neural rendering through quantum circuits, as shown in Figure 2. Additionally, we present a quantum volume rendering approach based on quantum integration, which has been successfully implemented on real quantum hardware. Our proposed quantum volume rendering performs the fastest among various rendering tasks and offers speed-related advantages over implementing conventional integration on digital computers. In brief, our contributions include:

(1) We introduce the first QNN-based system capable of rendering photorealistic novel views at speeds hundreds of times faster than conventional neural networks.

(2) The QRF framework is presented, comprising a collection of quantum implicit fields. For each quantum circuit, encoding circuits are learned to capture local properties, enabling high-quality rendering.

(3) We harness the power of the quantum activation function, which excels at representing signal details compared to classical activation functions in implicit

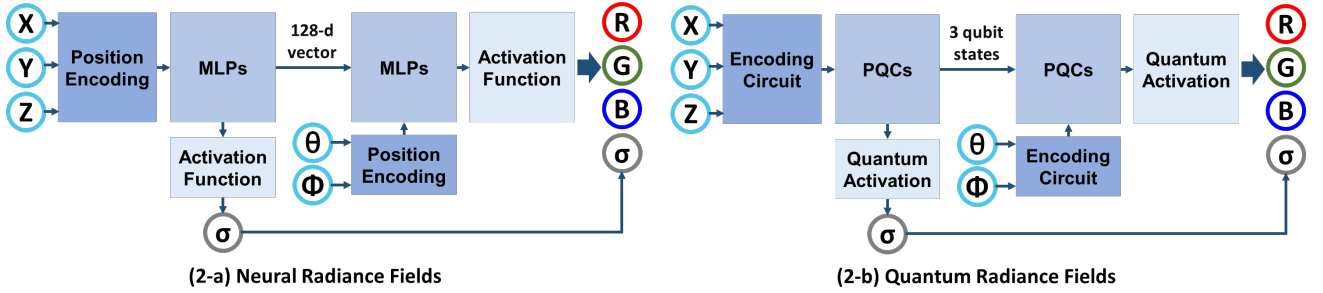

Figure 2: (2-a) NeRF architecture. Given a 3D position (x, y, z), viewing direction ($\theta, \phi$), NeRF produces static and transient colors (r, g, b) and transparency values ($\sigma$). (2-b): our QRF architecture replaces the same task with encoding circuit, parameterized quantum circuits, and quantum activation.

neural representations.

(4) We put forth Quantum Volume Rendering, a quantum algorithm for numerical integration, which fundamentally outperforms classical Monte Carlo integration in the context of volume rendering.

The remainder of this paper is organized as follows. In Section 2, we delve into the realm of neural scene representations, neural radiance fields, and QNN. Section 3 takes us into the domain of implicit neural representations, guided by QRF. In Section 4, we unveil our research outcomes through the lens of 2D image regression and 3D scene reconstruction. Within each task, we illustrate the advantages arising from the utilization of circuit-based QRF. Lastly, Section 5 encapsulates our study's essence, offering both a concise summary and a glimpse into potential future avenues of exploration.

## 2. Related Work

**Neural Scene Representation**. To model objects in a scene, many different scene geometry representations have been proposed. They can be divided into explicit and implicit representations. Explicit scene representations portray scenes as an assembly of geometric primitives, with their output falling into voxel-based [1,2], point-based [3,4], and mesh-based representations [5,6]. While explicit scene representations allow for the swift new perspectives generation, they are inherently constrained by the internal resolution of their representations, often resulting in blurry outcomes when dealing with high-frequency content.

To address the aforementioned challenge, numerous studies have delved into the realm of implicit neural scene representations, which enable direct inference of outputs from a continuous input space. In contrast to explicit neural scene representations, implicit neural scene representations offer the potential for 3D structure-aware, continuous, memory-efficient portrayals of shape components, objects, or entire scenes [7, 8, 9]. These representations employ neural networks to implicitly define objects or scenes and can be supervised directly using 3D data, such as point clouds, or through 2D multi-view images [10,11,12].

**Neural Radiance Fields**. Recent advancements in Neural Radiance Fields (NeRF) have demonstrated the capability of neural networks to acquire an implicit volumetric representation of a scene. They enable the encoding of intricate 3D environments that can be realistically rendered from novel perspectives [13]. Nevertheless, it is important to note that NeRF requires extensive point sampling along the ray to accumulate colors for achieving high-quality rendering.

Numerous techniques have emerged to accelerate NeRF. Neural Sparse Voxel Fields (NSVF) [15] utilize classical approaches such as empty space skipping and early ray termination to expedite NeRF's rendering. MetaNeRF [16] adopts standard meta-learning methods to learn initial weight parameters, leading to quicker convergence and improved reconstruction quality during test-time optimization. AutoInt [17] introduces an automatic integration framework with closed-form integral solutions, reducing the need for ray evaluations during NeRF's raymarching. DONeRF [18] enhances inference by decreasing the required number of samples along the ray. FastNeRF [19] proposes a graphics-inspired factorization technique that efficiently caches and queries data to compute pixel values in rendered images. KiloNeRF [20] represents scenes using numerous smaller MLP instead of a single large-capacity MLP, facilitating the use of smaller and faster evaluation MLP. DS-NeRF [21] employs depth as an additional source of supervision to regularize NeRF's geometry learning and improve training. Cheng Sun and Min Sun et al. [22] present a super-fast convergence method for reconstructing per-scene radiance fields, drastically reducing training time from hours to just 15 minutes while maintaining quality on par with NeRF.

**Quantum Neural Networks**. QNN represent a nascent domain that amalgamates the computational advantages introduced by quantum computing with advancements beyond classical computation [23]. These networks not only enhance algorithmic efficiency but also exhibit a heightened probability of converging to the global minimum when seeking solutions [24]. At their core, quantum computing principles draw inspiration from quantum physics, encompassing concepts like superposition, entanglement, and interference. A qubit system has the capacity to hold multiple bits of information concurrently, thus unlocking the potential for extensive parallelism [25].

## 3. Method

To achieve real-time, cinematic-quality rendering of concise neural representations for generated content, we adopt a QNN-based neural raymarching approach inspired by NeRF [13], which employs MLPs to encode densities and colors at continuous 3D positions within the scene. Drawing inspiration from DONeRF [18], we adopt a compact local sampling strategy for scene representation. This strategy enables the raymarching-based neural representation to focus on essential samples around the surface region, thereby reducing the utilization of qubits in simulated environments and quantum computers. In the upcoming sections, we provide an overview of the NeRF baseline model in Sec. 3.1, detail our QRF in Sec. 3.2, introduce the Quantum Activation Function in Sec. 3.3, and present Quantum Volume Rendering in Sec. 3.4.

### 3.1. Baseline Model

In the NeRF, the scene representation is accomplished by a neural network, denoted as $f_\theta$ with parameter $\theta_n$, designed to capture the 3D volumetric information. NeRF's neural network, $f_\theta: (p, d) \to (c, \sigma)$, maps a 3D position $p \in \mathbb{R}^3$ and a light direction $d \in \mathbb{R}^2$ to the color value $c$ and transparency $\sigma$. The architecture of $f_\theta$ is thoughtfully structured to ensure that only the color $c$ relies on the viewing direction $d$. This design choice promotes the learning of consistent scene geometry. In the preliminary deterministic preprocessing step, both $x$ and $d$ undergo a transformation via positional encoding, represented as $\gamma$. This encoding strategy facilitates the learning of high-frequency details in the data.

To render an individual image pixel, a ray is projected from the camera's center through the pixel and into the scene, with the direction of this ray denoted as $d$. Multiple 3D positions $(p_1, ... p_k)$ are sampled along the ray between the near and far boundaries defined by the camera parameters. At each position $p_i$ along the ray and for the given ray direction $d$, the neural network $f_\theta$ is evaluated, producing color $c_i$ and transparency $\sigma$ as intermediate outputs. These intermediate results are subsequently integrated to yield the final pixel color $\hat{c}$ as follows:

$$\hat{c} = \sum_i^K T_i(1 - e^{(-\sigma_i \delta_i)}) c_i \quad (1)$$

where $T_i = e^{-\sum_{j=i}^{i-1} \sigma_j \delta_j}$ is the transmittance and $\delta_j = (p_{i+1} - p_i)$ is the distance between samples. As $f_\theta$ is contingent on the ray direction, NeRF has the capacity to capture viewpoint-dependent effects, including specular reflections. This capability marks a significant dimension in which NeRF outperforms traditional 3D representation.

### 3.2. Quantum Radiance Fields

The QRF is implemented by a series of quantum circuits built into the continuous-variable architecture. It consists of three consecutive parts (as shown in Figure 2). An encoding circuit encodes the classical data into the states of the qubits, followed by a parameterized quantum circuits (PQCs), which is used to transform these states to their optimal location on the Hilbert space. Lastly, quantum activation is employed to perform nonlinear mapping on the input, introducing nonlinear factors into the neural networks. This augmentation enables the neural networks to more effectively capture the color and transparency attributes at each position along the ray.

#### 3.2.1 Encoding Circuit

The encoding circuit is used to encode the classical data into the physical states of Hilbert space for quantum computing [26], which is critical to the success of QNN. Quantum encoding can be thought of as loading a data point $x \in X$ from memory into a quantum state so that it can be processed by a QNN. The loading is accomplished by encoding from the set $X$ to the $n$-qubit quantum state $D_n$. Numerous QNN papers [27, 28, 29] have advocated wavefunction encoding with $n = log_2 2N$. This approach offers substantial space savings, albeit at the expense of a significant increase in computational time. In essence, a quantum state composed of $log_2 2N$ qubits can represent a data point with $N$ features, but the preparation of such a quantum state demands $O(2^n)$ time. Recent researchers [30, 31, 32] have explored alternative methods, such as angle encoding, which efficiently encodes classical data into quantum states, effectively mitigating the time complexity concerns associated with wave encoding.

**Angle Encoding**. Angle encoding makes use of rotation gates to encode classical information $x_k \in \mathbb{R}^N$ without any normalization condition. Angle encoding can be constructed using a single rotation with angle $\theta_k$ (normalized to be in $[-\pi, \pi]$) for each qubit, and can therefore encode $N$ features with $N$ qubits. Angle encoding consists in the following transformation:

$$S_k |0\rangle = \otimes_{k=0}^N cos(\theta_k) |0\rangle + sin(\theta_k) |1\rangle \quad (2)$$

where the circuit starts with the $|0\rangle$ state and proceeds to encode a data point $x_k$ using a circuit $S_k$. The number of qubits, denoted as $N$, equals the dimension of the vector $x_k$ that is being encoded. Angle encoding is remarkably straightforward and characterized by a depth of only 1. The primary advantage of angle encoding lies in its operational efficiency: It requires only a constant number of parallel operations, regardless of the quantity of data values that need to be encoded. However, from a qubit perspective, this approach is not optimal since each component of the input vector necessitates a dedicated qubit [40].

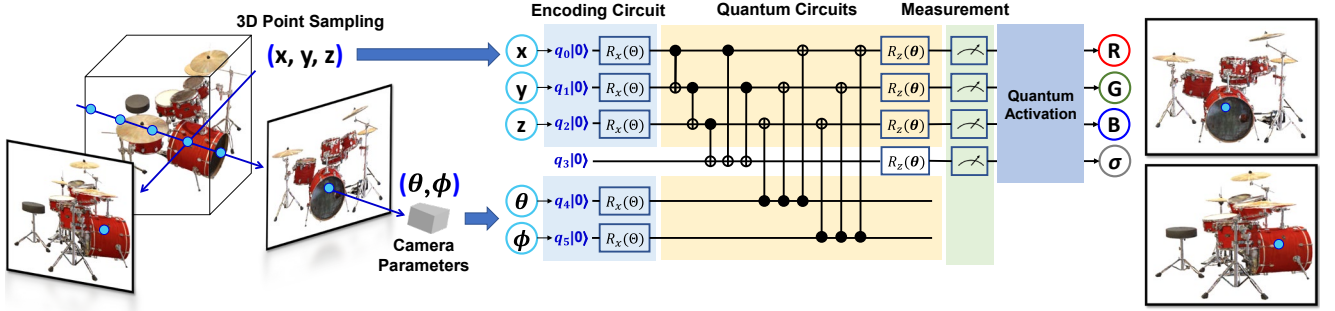

Figure 3: Quantum Radiance Fields with encoding circuits and quantum circuits produces colors (r, g, b) and transparency values ($\sigma$) given a 3D position (x, y, z) and viewing direction ($\theta$, $\phi$). Similar to the NeRF architecture, QRF enforces that the predicted $\sigma$ is independent of view direction. Note that this schematic is a simplified quantum circuit with only 4 rotation gates around the z axis.

**Dense Angle Encoding**. Angle encoding can be slightly generalized to encode two features per qubit by exploiting the relative phase degree of freedom [41]. We refer to this as the dense angle encoding and include a definition below:

$$|x\rangle = \otimes_{k=1}^{N/2} \cos(\theta_{2k}) |0\rangle + e^{2\pi i x_{2k}} \sin(\pi \theta_{2k}) |1\rangle \quad (3)$$

where $x$ is a feature vector $x = [x_1, \ldots, x_N]^T \in \mathbb{R}^N$, the dense angle encoding maps $x \to E(x)$. Dense encoding is derived by extending the above formula into two features using relative phase degrees of freedom. It exploits the additional property of relative phase qubits to encode $N$ data points using only $N/2$ qubits.

### 3.2.2 Parametrized Quantum Circuits.

Parametrized Quantum Circuit is composed of a set of parameterized single and controlled single qubit gates. The parameters are iteratively optimized by a classical optimizer to attain a desired input-output relationship. A block-diagonal approximation to the Fubini-Study metric tensor of a PQCs can be evaluated on quantum hardware. In general, an $n$ qubits PQCs can be written as:

$$u(\hat{\theta})|\varphi_0\rangle = \left(\prod_{\ell=1}^{k} W_\ell u_\ell(\theta_\ell)\right)|\varphi_0\rangle \quad (4)$$

where $\varphi_0$ is the initial quantum state, m is the maximum circuit depth, $W_\ell$ is the non-parameterized quantum gate at $\ell$-th layer, $u_\ell(\theta_\ell)$ is the parametrized quantum gate with parameters $\{\theta_0, \theta_1, \ldots \theta_k\}$ at $\ell$-th layer, which is a sequence consisting of parameterized qubit gates. Herein, the form of $u_\ell(\theta_\ell)$ is variable and accords with any physical constraint such as highly limited connectivity between physical qubits.

To achieve better entanglement of the qubits before appending nonlinear operations, the n qubits PQCs has n repeated layers in our model. In order to provide computational speedup by orchestrating constructive and destructive interference of the amplitudes in quantum computing, we constructed $m$ rotation gates on the $n$ qubits PQCs as our basic quantum circuit, which can be written as:

$$\left(\prod_{\ell=1}^{n} \left(\otimes_{j=0}^{m} CNOT_{i,i+1} R(\theta_{i+n \times j})\right)\right) \quad (5)$$

where $CNOT_{i,i+1}$ represents $CNOT$ gate as the control qubit. $R(\theta_{i+n \times j})$ represents the rotation gate along each of the X, Y, and Z-axis. $\theta_{i+n \times j}$ is adjustable parameter of rotation gates $R$. With the Pauli matrix, we can define single-qubit rotation along each of the X, Y, and Z-axis as:

$$R_x(\theta) = e^{-i\frac{\theta}{2}\sigma_x} = \begin{bmatrix} \cos(\theta/2) & -i\sin(\theta/2) \\ -i\sin(\theta/2) & \cos(\theta/2) \end{bmatrix} \quad (6)$$

$$R_y(\theta) = e^{-i\frac{\theta}{2}\sigma_y} = \begin{bmatrix} \cos(\theta/2) & -\sin(\theta/2) \\ \sin(\theta/2) & \cos(\theta/2) \end{bmatrix} \quad (7)$$

$$R_z(\theta) = e^{-i\frac{\theta}{2}\sigma_z} = \begin{bmatrix} e^{-i\frac{\theta}{2}} & 0 \\ 0 & e^{i\frac{\theta}{2}} \end{bmatrix} \quad (8)$$

where $\{\sigma_x, \sigma_y, \sigma_z\}$ is Pauli matrices. The operation of $R(\theta)$ can be modified by changing parameters $\theta$. Thus, the output state can be optimized to approximate the wanted state. By optimizing the parameters, the general PQCs tries to approximate arbitrary states so that it can be used for different specific molecules. The goal of PQCs is to solve an optimization problem encoded into a cost function:

$$\theta^* = \underset{\theta \in C}{argmin}(\langle \psi(\theta)|H|\psi(\theta)\rangle) \quad (9)$$

where $H$ is the Hamiltonian with the ground energy to seek. As parameters $\theta$ are continuous, many gradient-based optimization algorithms can be used to find the optimal ones. Figure 3 shows an example of Parametrized Quantum Circuit with $n$=6 and $m$=4. Four qubits use the rotation gate $R(\theta)$ by the angle $\theta$ around z-axis on the Hilbert space, and $CNOT$ gate is used for 2 specific qubits.

## 3.3. Quantum Activation Function

Recent implicit neural representations are built on ReLU-based multilayer perceptron. These architectures lack the capacity to represent the fine details in the underlying signal, and they often do not represent the derivative of the target signal well. This is partly due to the fact that ReLU networks are piecewise linear, and their second derivatives are zero everywhere, so they cannot model the information contained in the higher-order derivatives of natural signals. To address these limitations, we leverage quantum activation functions, which can better represent details in signals than ReLU-MLP for implicit neural representations.

We apply a multi-step quantum approach by selecting the ReLU's solution for positive values $R(z)_{ReLU}$, and the LReLU's solution for negative values $R(z)_{LReLU}$ [42]. By applying the quantum principle of entanglement, the tensor product of the two candidate Hilbert state spaces from $H_{ReLU}$ and $H_{LReLU}$ was performed as:

$$H_{ReLU} \otimes H_{LReLU} \quad (10)$$

The quantum entanglement in Eq. 10 allows to overcome the limitation of the ReLU being dying for negative inputs. The resulting state in the blended system is described by:

$$|\varphi\rangle_{ReLU} \otimes |\varphi\rangle_{LReLU} \quad (11)$$

In an entangled or inseparable state, the formulation of product states of Quantum ReLU (QReLU) can be generalized as:

$$|\varphi\rangle_{QReLU} = \sum_{ReLU,LReLU} |0|1\rangle_{ReLU} \otimes |0|1\rangle_{LReLU} \quad (12)$$

where keeping output for positive values in the QReLU, but with the added novelty of the entangled solution for negative values. This fits complicated signals, such as natural images and 3D shapes, and their derivatives robustly.

## 3.4. Quantum Volume Rendering

Conventional implicit neural representations represent a scene as an implicit function $F_\theta(p,v) \to (c,\omega)$, where $\theta$ are parameters of an underlying neural network [13]. It evaluate a volume rendering integral to compute the color of camera ray $p(z) = p_0 + z \cdot v$ as:

$$c(p_0, v) = \int_0^\infty \omega(p(z)) \cdot c(p(z), v) dz \quad (13)$$

where $\int_0^\infty \omega(p(z))dz = 1$, $c$ is the scene color, $w$ is the probability density at spatial location $p$ and ray direction $v$. $c(p_0, v)$ describes the scene color $c$ and its probability density $\omega$ at spatial location $p$ and ray direction $v$.

Volume rendering methods estimate the integral $c(p_0, v)$ by densely sampling points on each camera ray and accumulating the colors and densities of the sampled points into a 2D image as:

$$c(p_0, v) \approx \sum_{i=n}^{N} (\prod_{j=1}^{i-1} \alpha(z_j, \triangle_j)) \cdot (1 - \alpha(z_i, \triangle_i)) \cdot c(p(z_i), v) \quad (14)$$

where $\alpha(z_i, \triangle_i) = \exp(-\sigma(p(z_i), v))$, and $\{\sigma(p(z_i))\}_{i=1}^N$ are the colors and the volume densities of the sampled points.

Although volume rendering offer unprecedented image quality, they are also extremely slow and memory inefficient. This is because volume rendering methods need to sample a large number of points along the rays for color accumulation to achieve high quality rendering. Previous works only considered non-empty voxels for raymarching and reduce the number of samples per ray [15, 18]. However, for scenes with high depth and complexity, this works will result in longer evaluation time and lower rendering quality.

Volume rendering is essentially a numerical integration problem in each pixel, which is commonly done by Monte Carlo integration on classical computers. In this paper, we propose the quantum ray tracing, a quantum algorithm for numerical integration that is fundamentally superior to classical Monte Carlo integration. Furthermore, we apply Grover's search [33] to design clever algorithms to take full advantage of quantum parallelism. Given a ray tracing oracle that implements the following transformations:

$$O_f(pixel, channel): \sum_{j=0}^{N-1} x_j |j\rangle \to \sum_{j=0}^{N-1} x_j |j\rangle |f(j)\rangle \quad (15)$$

where $pixel$ and $channel$ (R, G or B) are classical parameters, $j$ plays the role of ray identity, and $f(j)$ is a real number that stands for the ray energy. In the rest of this paper $f$ is specified as the function that maps ray $j$ to ray energy. The oracle can trace $N = 2n$ paths simultaneously, and the final color we hope to write to the corresponding pixel and channel is the average of those energies $\frac{1}{N}\sum_{j=0}^{N-1} f(j)$. Suppose those real numbers $f(j)$ are stored in a fixed-point format with integer bit length $b_0$ and total bit length $b$, we can transfer the estimation problem of Eq. 15 into quantum counting by constructing a Boolean function:

$$g(j,k) = \begin{cases} 1, f(j) \geq 2^{b_0 - b_k} \\ 0, f(j) < 2^{b_0 - b_k} \end{cases} \quad (16)$$

where $k = (0,1,...,2^b - 1)$. The phase oracle $O_g$ for $g$ in Grover's search as:

$$O_g : \sum_{j,k} |j\rangle|k\rangle \to \sum_{j,k} (-1)^{g(j,k)} |j\rangle|k\rangle \quad (17)$$

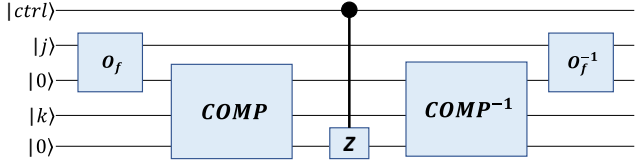

Figure 4: The construction of controlled-$O_g$

To construct $O_g$, we need a comparison gate that performs the comparison operation $COMP$ on the two integers $2^{b-b_0}f(j)$ and $k$,

$$COMP: \sum_{j,k}|f(j)\rangle|k\rangle \to \sum_{j,k}|f(j)\rangle|k\rangle|g(j,k)\rangle \quad (18)$$

The $O_g$ gate can be constructed as Figure 4. The quantity $\sum_{j,k}g(j,k)$ can be estimated by quantum counting algorithm. In the paper we assume one call to $O_f$ quantum ray tracing takes the same samples as tracing one path in classical numerical integration. We evaluate the cost of classical path tracing by the number of ray paths $N_c$, as the noise comes mostly from the Monte Carlo integration. And in quantum ray tracing, the time cost is evaluated by the number of queries $N_q$ to the ray tracing oracle $O_f$. The quantum integration has a convergence rate of $O(1/N_q)$, hence has a quadratic speedup over classical Monte Carlo integration with convergence rate of $O(1/\sqrt{N_c})$.

## 4. Results
### 4.1. Task

**2D Image Regression**. We train a QNN to regress from 2D input pixel coordinates to the corresponding RGB values of an image (as shown in Figure 5). We consider two different distributions $\mathcal{H}$: face images (CelebA [34]) and natural images (Div2K [35]). Given a sampled image $\hbar \sim \mathcal{H}$, we resize all images to 256 × 256 as observations for network weights $\theta$ in the optimization inner loop. At each inner loop step, the entire image is reconstructed and used to compute the loss. We then compare the classical MLP and QNN over these two distributions.

**3D Scene Representation**. The goal of 3D scene representation is to generate a novel view of the scene from a set of reference images. We validate our QRF through an extensive series of ablation studies and comparisons to recent techniques for accelerating NeRF. We evaluate our method on three inward-facing datasets:

(1) Synthetic-NeRF [13]: The Synthetic-NeRF dataset consists of 360-degree views of complex objects in 8 scenes, where each scene has a central object with 100 inward facing cameras distributed randomly on the upper hemisphere. The images are 800×800 with provided ground truth camera poses.

(2) Synthetic-NSVF [15]: The Synthetic-NSVF contains 8 objects synthesized by NSVF. Strictly following the settings of NSVF, we set the image resolution to 800 × 800 pixels and let each scene have 100 views for training and 200 views for testing.

(3) Tanks & Temples [36]: The Tanks and Temples is a real-world dataset containing 5 scenes of real objects captured by an inward-facing camera surrounding the scene. Each scene contains between 152-384 images of size 1920 × 1080.

### 4.2. Implementation

The principal baseline for our experiments is NeRF [13]. We report the results of the original NeRF implementation, as well as the reimplementation in Jax (JaxNeRF) [14]. We also compare two older methods, Scene Representation Network (SRN) [7] and Neural Volume [8], as well as five recent papers introducing NeRF accelerations, Neural Sparse Voxel Field (NSVF) [15], AutoInt [17] ], FastNeRF [19], KiloNeRF [20] and Depth-supervised NeRF (DS-NeRF) [21]. To evaluate QRF, we focus on two competing requirements of scene representations: quality of the generated images, and efficiency of the image generation. To quantify the rendering quality, we rely on three metrics:

(1) Peak Signal to Noise Ratio (PSNR): A classic metric to measure the corruption of a signal.

(2) Structural Similarity Index Measure (SSIM) [37]: A perceptual image quality assessment based on the degradation of structural information.

(3) Learned Perceptual Image Patch Similarity (LPIPS) [38]: A perceptual metric based on the deep features of a trained network that is more consistent with human judgement.

where higher PSNR, SSIM and lower LPIPS is most desirable. We train both the NeRF baseline and QRF for a high number of iterations to find the limits of the representation capabilities of the respective architectures. We train each model for 350k iterations. For the classical scene representation, we train it in 48 hours using 2 Tesla V100 GPUs. For our 2D quantum representations and QRF model, we train it with Borealis, a photonic quantum processor from Xanadu that can be programmed and entangled. The inference time performance is measured on a Tesla V100 for classical NeRF and a Borealis for QRF.

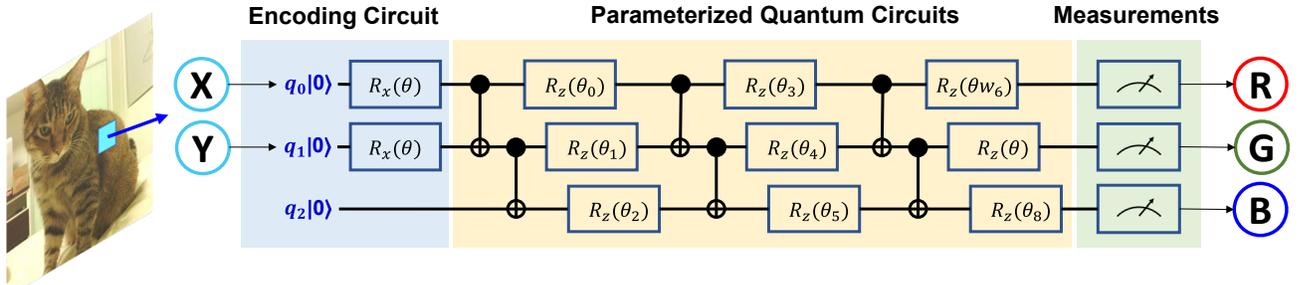

Figure 5: Quantum Implicit Neural Representations on 2D image regression with Encoding Circuit and PQS (n=3, m=3)

### 4.3. Ablation Study

We use Div2K dataset [35] and the Drums scene of Synthetic-NeRF [13] to conduct the ablation experiments. The strategy of Activation Function, Encoding Circuit, and Quantum Circuit are discussed in this section.

**Activation Function**. We first compare the performance of the classical activation function and the quantum activation function in the 2D image regression and 3D scene representation task. We use Rectified Linear Unit (ReLU), Exponential Linear Unit (ELU), Smooth ReLU (SoftPlus), SIREN [9] as the classical activation function, and QReLU as quantum activation function. For 2D image regression, we train an MLP with 4 layers/256 channels and apply sigmoid activation to the output for each task. For 3D scene representation, we train baseline NeRF with 6 layers/256 channels and apply positional encoding to the input coordinates. Table 1 shows that when the QRelu function serves as the activation function for each task, it outperforms the state-of-the-art activation function, which indicates that the quantum activation function has better convergence.

Table 1: Ablation Study of activation function from the Div2K and Synthetic-NeRF dataset.

| Task | | 2D Image Regression | | | 3D Scene Representation | | |
|---|---|---|---|---|---|---|---|
| Datasets | | Div2K Dataset [35] | | | Synthetic-NeRF [13] | | |
| Evaluate Metrics | | PSNR | SSIM | LPIPS | PSNR | SSIM | LPIPS |
| Classical AF | ReLU | 32.89 | 0.961 | 0.044 | 24.85 | 0.812 | 0.208 |
| | ELU | 23.12 | 0.380 | 0.258 | - | - | - |
| | Softplus | 19.37 | 0.273 | 0.482 | - | - | - |
| | SIREN | 36.92 | 0.970 | 0.021 | 26.44 | 0.907 | 0.179 |
| Quantum AF | QReLU | 38.43 | 0.971 | 0.016 | 27.12 | 0.919 | 0.168 |

**Encoding Circuit and Quantum Circuit**. We validate our QRF on our quantum system using various encoding circuits and quantum circuits. We use circuit-5/6/16/17 provided by Sukin Sim et al. [39], which has better expressive ability and entanglement ability, as the quantum circuit baseline. In addition, we adopt general qubit encoding, wavefunction encoding, angle encoding, and dense angle encoding as our encoding circuit strategy. More details about our quantum circuit and encoding circuit can be found in the supplemental.

The ablation results yield many significant findings (as show in Table 2). First, circuit-5 and circuit-6 are the fully connected graph arrangement of qubits which led to both favorable expressibility and entangling capability. Therefore, the network model with circuit-5 and circuit-6 has a high PSNR regardless of the encoding circuit used. Second, circuit-5 and circuit-16 with controlled Z-rotation (CRz) gates outperform circuit-6 and circuit-17, respectively. This is because the CRz operations in the entangling block commute with each other and thus the effective unitary operation comprised of CRz gates can be expressed using unique generator terms that are fewer than the number of parameters for these gates.

Table 2: Ablation Study of encoding circuit and quantum circuit from the Synthetic-NeRF dataset.

| Evaluate Metrics | Encoding Circuit | Quantum Circuit | | | |
|---|---|---|---|---|---|
| | | 5 | 6 | 16 | 17 |
| PSNR | General Qubit Encoding | 26.52 | 26.33 | 25.52 | 24.82 |
| | Wavefunction Encoding | 26.58 | 26.39 | 25.58 | 24.88 |
| | Angle Encoding | 26.98 | 26.80 | 25.99 | 25.29 |
| | Dense Angle Encoding | 27.33 | 27.15 | 26.33 | 25.64 |
| SSIM | General Qubit Encoding | 0.893 | 0.875 | 0.833 | 0.809 |
| | Wavefunction Encoding | 0.880 | 0.885 | 0.838 | 0.820 |
| | Angle Encoding | 0.911 | 0.900 | 0.860 | 0.832 |
| | Dense Angle Encoding | 0.921 | 0.915 | 0.873 | 0.838 |
| LPIPS | General Qubit Encoding | 0.183 | 0.183 | 0.195 | 0.211 |
| | Wavefunction Encoding | 0.182 | 0.189 | 0.195 | 0.208 |
| | Angle Encoding | 0.175 | 0.176 | 0.188 | 0.206 |
| | Dense Angle Encoding | 0.169 | 0.176 | 0.183 | 0.199 |

### 4.4. Experiment Results

**2D Image Regression**. We first compare the performance between classical MLP and QNN model in the task of 2D image regression. According to the ablation study result, we use QReLU as the activation function, dense angle encoding as the encoding circuit, and circuit-5 as our parameterized quantum circuit. Table 3 shows that our proposed quantum model has significant advantages in each evaluate metric.

**3D Scene Representation.** We leverage the same approach (utilizing QReLU, dense angle encoding, and circuit-5) as QNN to construct the QRF architecture. As illustrated in Table 4, our findings demonstrate that QRF inference outperforms NeRF by a staggering factor of over 2000, even surpassing most alternative methods, with the exception of FastNeRF. This remarkable acceleration in QRF's inference speed can be attributed to its efficient caching and subsequent query computation for pixel values, achieved through a factorization approach. Moreover, QRF excels in all image quality metrics, making it a standout performer. Our method does not directly tackle the issue of training speed. We propose addressing the training speed of NeRF models by exploring techniques such as initializing through Meta Learning [16] or reconstructing the per-scene radiance field via DVGO NeRF [22].

Table 3: Quantitative comparisons for 2D image regression. Compared with classical MLP, our proposed QNN model outperforms in the rendering quality.

| Datasets | CelebA [33] | | | Div2K Dataset [34] | | |
|---|---|---|---|---|---|---|
| Evaluate Metrics | PSNR | SSIM | LPIPS | PSNR | SSIM | LPIPS |
| Classical MLP | 30.67 | 0.945 | 0.112 | 32.89 | 0.961 | 0.044 |
| QNN | 33.71 | 0.963 | 0.038 | 36.36 | 0.969 | 0.027 |

## 5. Conclusion

In this paper, we introduced Quantum Radiance Fields, an innovative extension to NeRF that leverages quantum computing to achieve photorealistic image rendering. Our approach not only significantly accelerates rendering but also yields higher-quality images. Furthermore, the acceleration strategy we presented for quantum computing may find broader applicability in other methods. We aim for this paper to serve as a demonstration of the potential of quantum computing in delivering effective solutions for scene representation and volume rendering.

Table 4: Quantitative results on each scene from the Synthetic-NeRF [13], Synthetic-NSVF [15], and Tanks and Temples [36]. We highlight the top 3 results in each column are color coded as Top 1, Top 2 and Top 3.

| Dataset | Synthetic-NeRF [13] | | | | Synthetic-NSVF [15] | | | | Tanks and Temples [36] | | | |
|---|---|---|---|---|---|---|---|---|---|---|---|---|
| Evaluate Metrics | PSNR | SSIM | LPIPS | FPS | PSNR | SSIM | LPIPS | FPS | PSNR | SSIM | LPIPS | FPS |
| SRN [7] | 22.26 | 0.846 | 0.170 | 0.909 | 24.33 | 0.882 | 0.141 | 1.304 | 24.10 | 0.847 | 0.251 | 0.250 |
| Neural Volumes [8] | 26.05 | 0.893 | 0.160 | 3.330 | 25.83 | 0.892 | 0.124 | 4.778 | 23.70 | 0.834 | 0.260 | 1.000 |
| NeRF [13] | 31.01 | 0.947 | 0.081 | 0.023 | 30.81 | 0.952 | 0.043 | 0.033 | 25.78 | 0.864 | 0.198 | 0.007 |
| JaxNeRF [14] | 31.69 | 0.953 | 0.049 | 0.045 | 31.49 | 0.958 | 0.026 | 0.065 | 27.94 | 0.904 | 0.168 | 0.013 |
| NSVF [15] | 31.75 | 0.953 | 0.047 | 0.815 | **35.18** | **0.979** | **0.015** | 0.095 | **28.42** | **0.907** | 0.153 | 0.163 |
| AutoInt [17] | 25.55 | 0.911 | 0.170 | 0.380 | 26.63 | 0.916 | 0.090 | 0.545 | 22.28 | 0.766 | 0.278 | 0.116 |
| DoNeRF [18] | **32.50** | **0.957** | **0.037** | 5.635 | 32.29 | 0.962 | 0.027 | 8.085 | 27.02 | 0.805 | 0.174 | 1.715 |
| FastNeRF [19] | 29.97 | 0.941 | 0.053 | **172.42** | 29.78 | 0.946 | 0.083 | **224.71** | 24.92 | 0.792 | 0.213 | **47.67** |
| KioNeRF [20] | 31.02 | 0.950 | 0.051 | 38.46 | 33.37 | 0.970 | 0.020 | 55.07 | 28.41 | **0.910** | **0.091** | 11.68 |
| QRF | **32.65** | **0.960** | **0.029** | **47.26** | **35.44** | **0.980** | **0.014** | **67.70** | **29.65** | 0.820 | **0.085** | **14.36** |

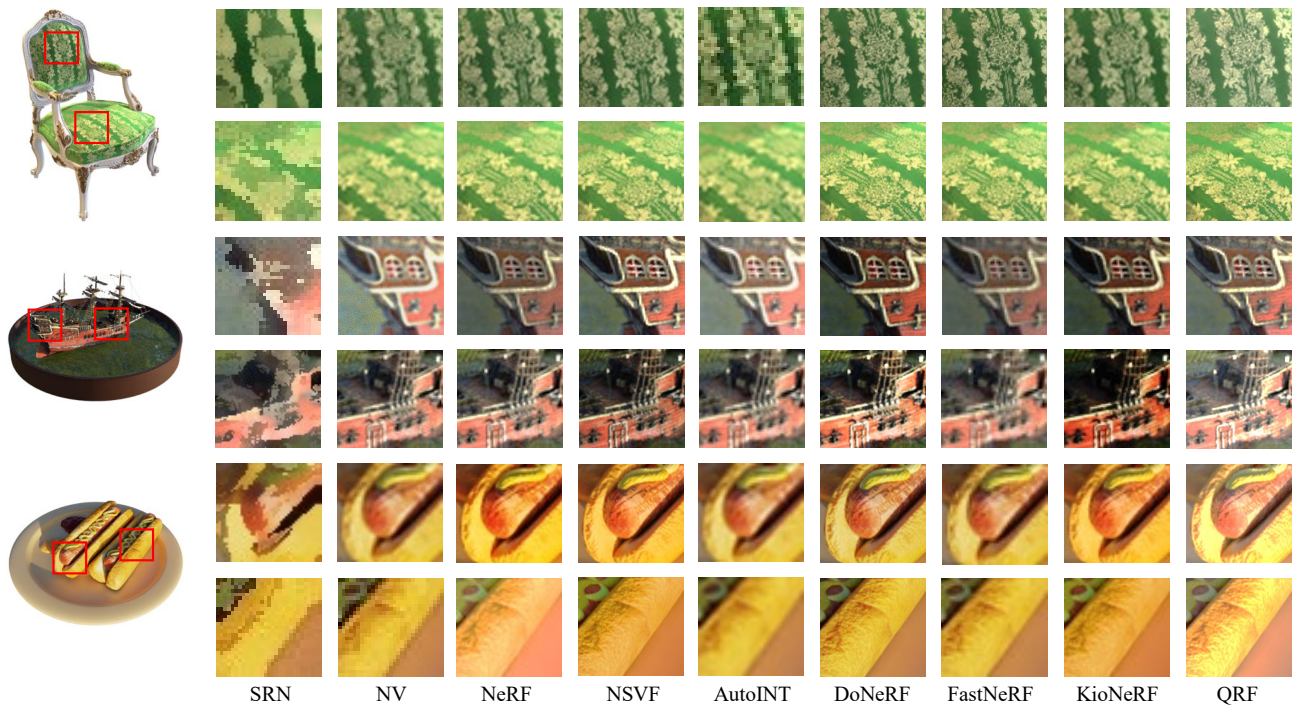

Figure 6: Qualitative comparisons on Synthetic-NeRF [13]. We compare classical implicit representation, NeRF, NeRF accelerations, and our proposed method. On this dataset, we find that our method better recovers the fine details in the scene. The results are similar in other datasets, please refer to our supplementary for more details.